\documentclass[letterpaper]{article} 
\usepackage{aaai23}  
\usepackage{times}  
\usepackage{helvet}  
\usepackage{courier}  
\usepackage[hyphens]{url}  
\usepackage{graphicx} 
\usepackage{natbib}  
\usepackage{caption} 
\usepackage{enumitem}
\usepackage{amsmath}
\usepackage{amssymb}
\usepackage{makecell}
\usepackage{multirow}

\usepackage{setspace}
\usepackage{verbatim}

\usepackage{algorithm}  
\usepackage{algorithmicx}  
\usepackage{algpseudocode}

\usepackage{latexsym}

\usepackage{subfigure}
\usepackage{color}

\usepackage{float}
\usepackage{newfloat}
\usepackage{listings}
\DeclareCaptionStyle{ruled}{labelfont=normalfont,labelsep=colon,strut=off} 
\floatstyle{ruled}
\newfloat{listing}{tb}{lst}{}
\floatname{listing}{Listing}

\title{DRGCN: Dynamic Evolving Initial Residual for Deep \\Graph Convolutional Networks}
\author{
    Lei Zhang\textsuperscript{\rm 1}\equalcontrib,
    Xiaodong Yan\textsuperscript{\rm 1}\equalcontrib,
    Jianshan He\textsuperscript{\rm 1},
    Ruopeng Li\textsuperscript{\rm 1},
    Wei Chu\textsuperscript{\rm 1}
}
\affiliations{
    \textsuperscript{\rm 1}Ant Group, Beijing, China\\

    \{xiaobei.zl, qiqing.yxd, yebai.hjs, ruopeng.lrp, weichu.cw\}@antgroup.com
}

\begin{document}

\maketitle

\begin{abstract}
Graph convolutional networks (GCNs) have been proved to be very practical to handle various graph-related tasks. It has attracted considerable research interest to study deep GCNs, due to their potential superior performance compared with shallow ones. However, simply increasing network depth will, on the contrary, hurt the performance due to the over-smoothing problem. Adding residual connection is proved to be effective for learning deep convolutional neural networks (deep CNNs), it is not trivial when applied to deep GCNs. Recent works proposed an initial residual mechanism that did alleviate the over-smoothing problem in deep GCNs. However, according to our study, their algorithms are quite sensitive to different datasets. In their setting, the personalization (dynamic) and correlation (evolving) of how residual applies are ignored. To this end, we propose a novel model called \textbf{D}ynamic evolving initial \textbf{R}esidual \textbf{G}raph \textbf{C}onvolutional \textbf{N}etwork (DRGCN). Firstly, we use a dynamic block for each node to adaptively fetch information from the initial representation. Secondly, we use an evolving block to model the residual evolving pattern between layers. Our experimental results show that our model effectively relieves the problem of over-smoothing in deep GCNs and outperforms the state-of-the-art (SOTA) methods on various benchmark datasets. Moreover, we develop a mini-batch version of DRGCN which can be applied to large-scale data.  Coupling with several fair training techniques, our model reaches new SOTA results on the large-scale \textit{ogbn-arxiv} dataset of Open Graph Benchmark (OGB)$\footnote{\url{ https://ogb.stanford.edu/docs/leader_nodeprop/#ogbn-arxiv }}$. Our reproducible code is available on GitHub$\footnote{\url{https://github.com/anonymousaabc/DRGCN}}$.
\end{abstract}

\section{Introduction}

    \begin{figure*}[t]
		\centering
		\includegraphics[width=0.9\textwidth]{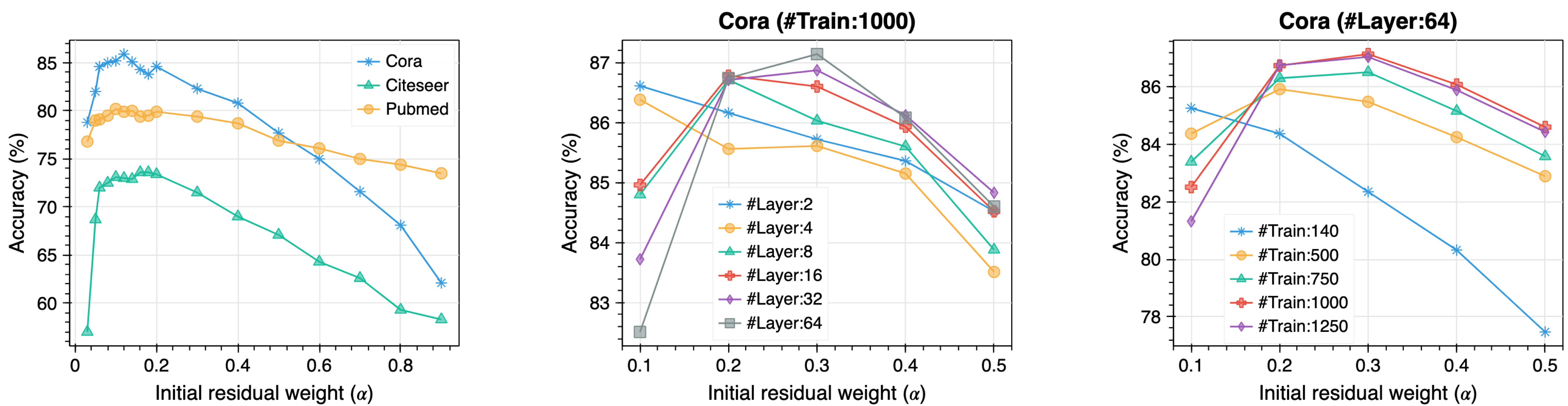}
		\caption{The sensitivity analysis results of fixed alpha for different datasets (left), different model depths (middle), and different training sizes (right).}
		\label{static_alpha}
	\end{figure*}

    Graph representation learning is the main task for graph-structured data. Graph Convolution Networks (GCNs) \cite{kipf2017semi} and its attentional variants (\textit{e.g.}, GAT) \cite{velivckovic2018gat} have been shown to be quite successful especially on graph representation learning tasks. These models are powerful on node classification \cite{kipf2017semi}, link prediction \cite{berg2017linkpred} and clustering \cite{chiang2019cluster} tasks, because of their remarkable ability to iteratively aggregate information from graph neighborhoods. GCNs also achieve great success on applications like recommender system \cite{ying2018pinsage,he2020lightgcn}. Motivated by deep neural networks from other areas such as computer vision \cite{he2016resnet}, where deeper models commonly have superior performance than shallow ones, researchers have made great progress on the design of deeper GCN structures. Intuitively, the deep version of GCNs has the ability to aggregate knowledge from remote hops of neighbors for target nodes. But it fails because these models end up with severe over-smoothing \cite{li2018deeper} issue, which shows that after too many rounds of the aggregation process, the representations of nodes converge to the same pattern and become indistinguishable.
	
	Several research efforts have been devoted to tackling the over-smoothing issue. Data augmentation strategies proposed from works like DropEdge \cite{rong2020dropedge} and GRAND \cite{feng2020grand} can be used as general techniques for various graph convolution networks. Recently, the residual connection is proved to be a powerful strategy to solve the over-smoothing issue. JKNet \cite{xu2018jknet} employs a dense skip connection to combine information of all layers for final node representation. APPNP \cite{klicpera2019appnp} proposes initial residual in the context of Personalized PageRank \cite{page1999pagerank} to constructs a skip connection to the initial representation. However, these models achieve their best results with shallow layers, and the performance drops when networks become deeper. GCNII \cite{chen2020gcnii} goes further with the initial residual by introducing identity mapping, and alleviates the over-smoothing issue while achieving new state-of-the-art results on various tasks.
	
	However, all of the previous works using residual connection treat residual weight as a fixed hyperparameter for all layers and nodes. The node personalization and layer correlation are ignored. Consequently, they are quite sensitive to datasets (Figure \ref{static_alpha}, left), the model depths (Figure \ref{static_alpha}, middle), and the scale of training sizes (Figure \ref{static_alpha}, right).
	
	Intuitively, because different nodes may have different initial representations and local topology information, personalizing residual weight for each node makes sense. Moreover, shallow layers still memory the information from the initial representation, which means we do not need to fetch much information by initial residual for them. On the other hand, as the layer goes deeper, we should fetch more information from the initial representation to prevent noise. Thus, we believe that initial residual weights from shallow to deeper layers should obey some evolving pattern. 
	
	Accordingly, we argue that a powerful deep residual GCN model should have the ability to learn the node personalized dynamic residual weight and uncover the residual evolving pattern from shallow to deep layers, so we introduce DRGCN.
    Firstly, we use a dynamic block for each node to adaptively fetch information from the initial representation. Secondly, we design an evolving block to capture the residual evolving pattern between layers. And we compare DRGCN with other deep residual GCN models on various datasets. The experimental results show that our model not only alleviates the over-smoothing issue but also outperforms the state-of-the-art models. 
    Furthermore, we visually study the learned residual weights, then we discover an evolving trend between shallow (short-distance) and deep (long-distance) layers, and the residual weight obeys some distribution characteristic, this verifies our intuition. 
    Finally, to improve the generalized ability of our model on large-scale datasets, we design the mini-batch version of DRGCN. 
    
    

    To summarize, the main contributions of our work are:
    
    \begin{itemize}
    \item We propose the dynamic evolving initial residual for deep GCN, which jointly consider node personalization and layer correlation. To the best of our knowledge, this is the first work introducing dynamic residual into GCNs to relieve the over-smoothing issue, and it could be a general technique for the deep GCN variants incorporating the residual connection.
    \item We conduct extensive experiments to demonstrate the consistent state-of-the-art performance of our model on various benchmark datasets. Our model also shows good self-adaption to different datasets, model depths, and scale of data sizes.
    \item We introduce mini-batch DRGCN which optimizes the memory consumption without degrading the performance. We reach new SOTA results on the large-scale ogbn-arxiv dataset of the OGB leaderboard \cite{hu2020ogb} at the time of submission (team anonymous), which demonstrates the generalization of our model on the large-scale dataset.
    \end{itemize}

    \begin{figure*}[t]
    \centering
	\includegraphics[width=0.85\textwidth]{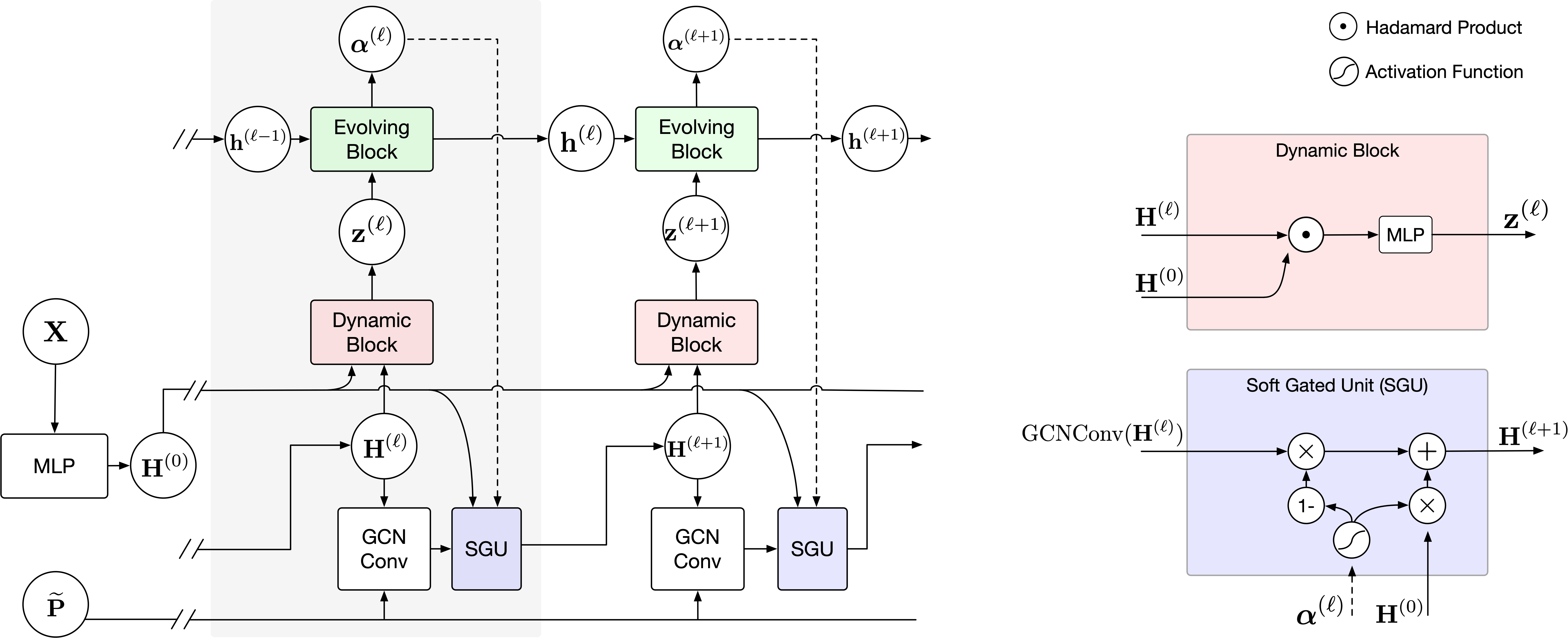}
    \caption{The overall architecture of DRGCN. Only layer $\ell$ and $\ell+1$ are shown for simplicity. Node features $\mathbf{X}$ and normalized adjacency matrix $\widetilde{\mathbf{P}}$ are inputs of this model. DRGCN consists of three components: 1) the Dynamic Block, which gets the node personalized initial residual $\mathbf{z}^{(\ell)}$ based on initial representation $\mathbf{H}^{(0)}$ and hidden representation $\mathbf{H}^{(\ell)}$; 2) the Evolving Block, which models the initial residual evolving pattern using $\mathbf{z}^{(\ell)}$ as input and produces $\boldsymbol{\alpha}^{(\ell)}$; 3) the SGU, which accepts $\mathbf{H}^{(0)}$, $\boldsymbol{\alpha}^{(\ell)}$, and the result of GCNConv as inputs and produces the final hidden representation of nodes.}
	
    \label{model_overall}
    \end{figure*}

\section{Preliminaries and Related Work}

In this section, we first define some essential notations used in this paper. Given a connected undirected graph $\mathcal{G}=(\mathcal{V},\mathcal{E})$ with $n$ nodes and $e$ edges, where $v_i\in\mathcal{V}$ and $(v_i,v_j)\in\mathcal{E}$
	denote node with index $i\in(1,2,\ldots,n)$ and edge between node $v_i$ and $v_j$ respectively. We use $\mathbf{X}\in\mathbb{R}^{n\times d}$ to denote the node feature matrix where $d$ is the feature dimension of the node. The topology information is described by the adjacency matrix $\mathbf{A}\in\mathbb{R}^{n\times n}$, and $\mathbf{A}_{ij}=1$ if $(v_i,v_j)\in\mathcal{E}$ , otherwise 0. For an undirected graph, $\mathbf{A}$ is a symmetric matrix. Let $\mathbf{D}$ denote the diagonal degree matrix, where $\mathbf{D}_{ii}=\sum_j\mathbf{A}_{ij}$ meaning the number of neighbors of $v_i$. When adding self-loop edges, the adjacency matrix and diagonal degree matrix become $\widetilde{\mathbf{A}}=\mathbf{A}+\mathbf{I}$ and $\widetilde{\mathbf{D}}=\mathbf{D}+\mathbf{I}$ respectively, with $\mathbf{I}$ being the identity matrix.

	\subsection{Deep GCNs}
	
	Recent researches mainly focus on two types of GCNs, spectral-based \cite{chen2020gcnii} and spatial-based \cite{shi2020unimp}. For a spectral-based manner, the vanilla GCN with two layers is expressed as 
	\begin{eqnarray}\label{vanilla_gcn}
		\mathbf{Z}=\widetilde{\mathbf{P}}\operatorname{ReLU}\left(\mathbf{\widetilde{\mathbf{P}}\mathbf{X}\mathbf{W}^{(0)}}\right)\mathbf{W}^{(1)},
	\end{eqnarray}
	where $\mathbf{Z}\in\mathbb{R}^{n\times c}$ is the output embedding matrix of target nodes and $c$ is the number of classes and $\widetilde{\mathbf{P}}=\widetilde{\mathbf{D}}^{-1/2}\widetilde{\mathbf{A}}\widetilde{\mathbf{D}}^{-1/2}$. In addition, $\mathbf{W}^{(0)}\in\mathbb{R}^{d\times d_h}$ and $\mathbf{W}^{(1)}\in\mathbb{R}^{d_h\times c}$ are weight matrices to map features. $d_h$ is the dimension of the hidden representation.
	
	To further improve the graph model performance, deep GCNs iteratively apply first-order graph convolution described in Eq. \ref{vanilla_gcn} vertically to aggregate information from the $(L-1)$-order neighborhood of target nodes. Here $L$ denotes the model depth, also the number of convolutional layers. So deep GCNs follow the same rule as
	\begin{eqnarray}\label{deep_gcn}
		\mathbf{H}^{(\ell+1)}=\sigma\left(\widetilde{\mathbf{P}}\mathbf{H}^{(\ell)}\mathbf{W}^{(\ell)}\right),
	\end{eqnarray}
	where $\sigma(\cdot)$ denotes the nonlinear activation function such as $\operatorname{ReLU}$. Meanwhile, $\mathbf{H}^{(\ell)}$ and $\mathbf{W}^{(\ell)}$ , $\ell\in(0,1,\ldots,L-1)$, are the input hidden node representation and weight matrix in the $\ell$-th layer. Specifically, $\mathbf{H}^{(0)}=\mathbf{X}$ and $\mathbf{H}^{(L)}$ is the node representation output. So for each graph convolution layer, the hidden node representations are updated through a pipeline of feature propagation, linear transformation, and point-wise nonlinear activation.
	
	DAGNN\cite{liu2020dagnn} demonstrates that the disentanglement of the transformation and propagation process improves the model performance. The transformation stage is defined as
	\begin{eqnarray}\label{init_h0}
		\mathbf{H}^{(0)}=\operatorname{MLP}(\mathbf{X}),
	\end{eqnarray}
	and the propagation stage is described as
	\begin{eqnarray}
		\mathbf{H}^{(\ell)}=\widetilde{\mathbf{P}}^{\ell}\mathbf{H}^{(0)},
	\end{eqnarray}
	where $\widetilde{\mathbf{P}}^{\ell}$ is the $\ell$-th power of $\widetilde{\mathbf{P}}$.

	\subsection{Deep Residual GCNs}
	
	The residual method has been demonstrated to be a simple but effective strategy to efficiently train very deep CNNs and improve their performance in the Computer Vision research area \cite{he2016resnet}. Motivated by this, another line of deep GCNs resort to residual methods to relieve the over-smoothing issue.
	
	ResGCNs \cite{li2019deepgcns} propose a dense residual connection that combines the smoothed representation $\widetilde{\mathbf{P}}\mathbf{H}^{(\ell)}$ with $\mathbf{H}^{(\ell)}$. And the output node representation for layer $\ell$ is
	\begin{eqnarray}\label{dense_res}
		\mathbf{H}^{(\ell+1)}=\sigma\left(\left(1-\alpha\right) \widetilde{\mathbf{P}} \mathbf{H}^{(\ell)}+\alpha \mathbf{H}^{(\ell)}\right),
	\end{eqnarray}
	where $\alpha$ is the residual weight hyperparameter for all nodes and layers. Recently, APPNP and GCNII propose the initial residual connection which combines $\mathbf{H}^{(\ell)}$ with the initial representation $\mathbf{H}^{(0)}$ rather than the node representation from the previous layer. The initial residual mechanism is able to receive information from the initial layer whose representation is proved to be crucial for nodes classification \cite{liu2020dagnn}. Formally, a deep GCN model with the initial residual  mechanism is defined as
	\begin{eqnarray}\label{initial_res}
		\mathbf{H}^{(\ell+1)}=\sigma\left(\left(1-\alpha\right) \widetilde{\mathbf{P}} \mathbf{H}^{(\ell)}+\alpha \mathbf{H}^{(0)}\right),
	\end{eqnarray}
	where the initial node representation $\mathbf{H}^{(0)}$ is obtained by $\mathbf{X}$ using Eq. \ref{init_h0}. Therefore, the initial residual connection ensures that the representation of each node in each layer retains at least $\alpha$ fraction information from the initial representation even if we stack large layers.
	
	Besides purely model design, another line of research mainly focuses on how to optimize the GPU memory consumption of deep GCNs in large-scale scenarios. RevGCN \cite{li2021gnn1000} proposes the reversible residual connection to increase the GPU memory efficiency and thus is able to train a deep GCN model with more than 1000 layers. Also, its variants show competitive results on large-scale graph datasets. We will compare our model with RevGCN in the large-scale scenario in Section \ref{large_scale_section}.

\section{DRGCN Model}

\subsection{Overall}
	
	In this section, we present the overall architecture of the DRGCN model. The proposed DRGCN framework is shown in Figure \ref{model_overall}.
	
	In summary, we introduce two core modules in DRGCN: 1) the Dynamic Block, which independently measures the similarity between initial and hidden representation, and dynamically determines the initial residual weight of each layer and node; 2) the Evolving Block, which models the initial residual evolving process sequentially and adaptively adjusts the weight of the initial representation in each layer. 
	
	We also propose a variant DRGCN* that additionally utilizes data augmentation. DRGCN* is more robust on datasets of different scales. We will describe the details of DRGCN* in section \ref{data_augmentation_section}. In the following, we will introduce the mathematical expressions of our model in detail.
	
	\subsection{The Dynamic Block}
	
	The top right of Figure \ref{model_overall} depicts the details of Dynamic Block. Formally, The mathematical expression is defined as
	\begin{eqnarray}\label{dirb_formula}
		\mathbf{z}^{(\ell)}=\operatorname{MLP}\left(\Phi\left(\widetilde{\mathbf{H}}^{(0)},\widetilde{\mathbf{H}}^{(\ell)}\right)\right),
	\end{eqnarray}
	where $\Phi$ denotes the combination function, $\widetilde{\mathbf{H}}^{(0)}\in\mathbb{R}^{n\times d_h}$ is obtained by L2 normalization from the first layer representation $\mathbf{H}^{(0)}$. Similarly, $\widetilde{\mathbf{H}}^{(\ell)}\in\mathbb{R}^{n\times d_h}$ is the L2 normalization of the $\ell$-th layer representation $\mathbf{H}^{(\ell)}$. The combination function generally uses Hadamard product operation. Nevertheless, $Sub$ and $Concat$ can also be used as $\Phi$. 
	
	Theoretically, MLP can approximate any measurable function \cite{hornik1989multilayer}. Therefore, for the purpose of modeling the similarity between $\mathbf{H}^{(0)}$ and $\mathbf{H}^{(\ell)}$, we use MLP to map the vector into residual weight scalar for each node. Hence, $\mathbf{z}^{(\ell)}\in\mathbb{R}^{n\times 1}$.
	
	\subsection{The Evolving Block}
	
	The dynamic block takes the initial and hidden representation of the current layer as inputs and doesn't take residual dependencies between layers into account. Therefore, $\mathbf{z}^{(\ell)}$ is not the final residual weight. We treat the residual dependencies modeling as a sequential evolving problem from shallow to deep layers, and we apply the evolving block to model the initial residual evolving process as
	\begin{eqnarray}\label{evolve_rnn_formula}
		\boldsymbol{\alpha}^{(\ell)}, \mathbf{h}^{(\ell)}=g\left(\mathbf{z}^{(\ell)}, \mathbf{h}^{(\ell-1)}\right),
	\end{eqnarray}
	 where $g$ is the evolving function and can be RNN, LSTM or their variants, $\mathbf{h}^{(\ell)}\in\mathbb{R}^{n\times1}$ is the hidden state output of the $\ell$-th layer. Specially, $\mathbf{h}^{(0)}$ is initialized from $\mathcal{U}(-\sqrt{k}, \sqrt{k})$ where $k=1/d_h$. $\boldsymbol{\alpha}^{(\ell)}\in\mathbb{R}^{n\times1}$ is the output of evolving block and also the final initial residual weight in the $\ell$-th layer. $\boldsymbol{\alpha}^{(\ell)}$ measures how much information of the initial representation should be retained in each propagation layer.
	 
	 \subsection{Propagation and Soft Gated Unit}
	 In propagation stage, we use the spectral convolution described in Eq. \ref{deep_gcn} expect that we use $\mathbf{W}^{(\ell)}$ only for the last layer. The propagation for aggregating messages in the $\ell$-th layer is expressed as $\widetilde{\mathbf{P}} \mathbf{H}^{(\ell)}$, which is displayed as ``GCNConv'' in Figure \ref{model_overall}. Remarkably, we can also use attentive based propagation methods, which form DRGAT for large-scale datasets, see section \ref{large_scale_section}.
	
	
	
	As is shown at the bottom right of Figure \ref{model_overall}. For the last component SGU, to narrow the value range of the residual weight, we utilize an activation function  $Sigmoid$ to $\boldsymbol{\alpha}^{(\ell)}$, and the final $\boldsymbol{\alpha}^{(\ell)}$ is layer-dependent and node-personalized. Then, we simply use a weight sum operation to combine the initial representation and hidden representation. Formally, the $\ell$-th layer output of the SGU module is defined as
	\begin{eqnarray}\label{sgu_formular}
		\mathbf{H}^{(\ell+1)}=\operatorname{ReLU}\left(\left(1-\boldsymbol{\alpha}^{(\ell)}\right) \widetilde{\mathbf{P}} \mathbf{H}^{(\ell)}+\boldsymbol{\alpha}^{(\ell)} \mathbf{H}^{(0)}\right).
	\end{eqnarray}

	And the final prediction result is 
	\begin{eqnarray}\label{drgcn_prediction}
		\widehat{\mathbf{Y}}=\operatorname{softmax}\left(\operatorname{MLP}\left(\mathbf{H}^{(L)}\right)\right).
	\end{eqnarray}
	
	\subsection{The Data Augmentation}\label{data_augmentation_section}
		
		To make DRGCN suitable for modeling data of various scales. We propose DRGCN* by referring to data augmentation from GRAND \cite{feng2020grand}. 
		
		Firstly, performing $S$ times of data augmentation, naturally, we get the prediction result of the $s$-th data augmentation as $\widehat{\mathbf{Y}}^{(s)}, s\in(1,2,\ldots,S)$, referring to Eq. \ref{drgcn_prediction}. Suppose that the true label is $\mathbf{Y}$, and the supervised loss of the node classification task is defined as
		\begin{eqnarray}\label{sup_loss}
			\mathcal{L}_{\text{sup}}=-\frac{1}{S} \sum_{s=1}^{S} \sum_{i=1}^{n} \mathbf{Y}_{i} \widehat{\mathbf{Y}}_{i}^{(s)}.
		\end{eqnarray}
		
		Secondly, we implement the consistency regularization loss to control the $S$ prediction results to be as same as possible. We calculate the distribution center of the prediction label by $\overline{\mathbf{Y}}_{i}=\frac{1}{S} \sum_{s=1}^{S} \widehat{\mathbf{Y}}_{i}^{(s)}$. Then we utilize the sharpening trick \cite{feng2020grand} to get the labels based on the average distributions $\overline{\mathbf{Y}}_{i} \rightarrow \overline{\mathbf{Y}}_{i}^{\prime}$. The consistency regularization loss $\mathcal{L}_{\text{con}}$ is obtained by minimizing the distance between $\overline{\mathbf{Y}}_{i}^{\prime}$ and $\mathbf{Y}_{i}^{(s)}$
		\begin{eqnarray}\label{con_loss}
			\mathcal{L}_{\text{con}}=\frac{1}{S} \sum_{s=1}^{S} \sum_{i=1}^{n}\|\overline{\mathbf{Y}}_{i}^{\prime}-\widehat{\mathbf{Y}}_{i}^{(s)}\|_{2}^{2}.
		\end{eqnarray}
		
	 	Finally, when using $\lambda$ to balance the two losses (as an usual case $\lambda=1$), the final loss of DRGCN* is defined as
		\begin{eqnarray}\label{all_loss}
			\mathcal{L}=\mathcal{L}_{\text{sup}}+\lambda\mathcal{L}_\text{con},
		\end{eqnarray}
		
		By introducing these two training techniques, the over-fitting issue of DRGCN on the small-scale dataset can be effectively solved. In addition, the computational complexity of DRGCN* is linear with the sum of node and edge counts \cite{feng2020grand}.

\section{Experiments}

    
    \subsection{Dataset and Experimental setup}\label{dataset_section}
    
    First, we use three standard citation network datasets Cora, Citeseer, and Pubmed \cite{sen2008collective} for semi-supervised node classification. 
    Then, we conduct the experiments on the Node Property Prediction of Open Graph Benchmark\cite{hu2020ogb}, which includes several various challenging and large-scale datasets. The statistics of overall datasets are summarized in Table \ref{data_summary}.

    In the experiments, we apply the standard fixed validation and testing split \cite{kipf2017semi} on three citation datasets with 500 nodes for validation and 1,000 nodes for testing. In addition, in training set sizes experiments, we conduct experiments with different training set sizes \{140,500,750,1000,1250\} on the Cora dataset for several representative baselines. Then, according to the performance of the training set sizes experiments, we adopted fixed training set sizes for model depths experiments,  which are 1000 for Cora, 1600 for Citeseer, and 10000 for Pubmed, and we aim to provide a rigorous and fair comparison between different models on each dataset by using the same dataset splits and training procedure. In addition, OGB dataset splitting directly calls the official interface.
    And, the results of all the experiments are averaged over 20 runs with random weight initializations.

    \addtolength{\tabcolsep}{-4pt}
    \begin{table}[t]
        \centering
        \caption{\label{data_summary}Summary of the datasets used in our experiments.}
        \begin{tabular}{l|c|c|c|c}
            \hline 
            \textbf{Dataset} & \textbf{Classes} & \textbf{Nodes} & \textbf{Edges} & \textbf{Features} \\
            
            \hline
            \textbf{Cora} & 7 & 2,708 & 5,429 & 1433 \\
            \textbf{Citeseer} & 6 & 3,327 & 4,732 & 3703  \\
            \textbf{Pubmed} & 3 & 1,9717 & 44,338 & 500 \\
            
            \hline
            \textbf{ogbn-arxiv} & 40 & 169,343 & 1,166,243 & 128 \\
            
            \textbf{ogbn-products} & 47 & 2,449,029 & 61,859,140 & 100 \\
            
            \hline
        \end{tabular}
    \end{table}
    \addtolength{\tabcolsep}{4pt}

    \subsection{Baseline Models}\label{basemodel_section}
    
    We compare our DRGCN with the following baselines:
    \begin{itemize}
    \item \textbf{GCN} \cite{kipf2017semi} which uses an efficient layer-wise propagation rule that is based on a first-order approximation of spectral.
    \item \textbf{GAT} \cite{velivckovic2018gat} which leverages masked self-attentional layers to address the shortcomings of prior methods based on graph convolutions or their approximations. 
    \item \textbf{DAGNN} \cite{liu2020dagnn} that incorporates message from large receptive fields through the disentanglement of representation transformation and propagation.
    \item \textbf{GRAND} \cite{feng2020grand} that uses the random propagation strategy and consistency regularization to improve the model's generalization.
    \item \textbf{GCNII} \cite{chen2020gcnii} that prevents over-smoothing by initial residual connection and identity mapping.
    \item \textbf{RevGAT} \cite{li2021gnn1000} which integrates reversible connections, group convolutions, weight tying, and equilibrium models to advance the memory and parameter efficiency for deep GCNs.
    \end{itemize}

    \addtolength{\tabcolsep}{-3pt}
    \begin{table}[h!]
        \centering
        \caption{\label{detail_result_summary}Summary of classification accuracy(\%) results with various depths. DRGCN* is the variant of DRGCN that additionally utilizes data augmentation.
        }
        \begin{tabular}{l|c|cccccc}
            \hline 
            \textbf{Dataset} & \textbf{Method} & \multicolumn{6}{c}{\textbf{Layers}} \\
            
             &  & 2 & 4 & 8 & 16 & 32 & 64 \\
            \hline
                    & GCN & 81.9 & 78.3 & 68.5 & 60.2 & 23.0 & 14.1\\
                    & GAT & 83.5 & 79.3 & 65.2 & 60.1 & 35.3 & 16.2 \\
                    & DAGNN & 68.2 & 83.0 & 87.2 & 87.3 & 86.6 & 84.6  \\
                    {\textbf{Cora}} & GRAND & 75.9 & 80.8 & 80.9 & 76.6 & 71.2 & 62.8  \\
                    & GCNII & 84.6 & 85.6 & 86.0 & 86.4 & 86.7 & 87.1  \\
            \cline{2-8}
                    & \textbf{DRGCN} & 85.9 & 86.7 & 87.0 & 87.3 & 87.3 & 87.5  \\
                    & \textbf{ DRGCN*} & \textbf{86.6} & \textbf{87.2} & \textbf{87.4} & \textbf{87.8} & \textbf{87.9} & \textbf{88.1} \\
            \hline
                    & GCN & 76.8 & 66.2 & 32.4 & 14.4 & 14.2 & 10.2\\
                    & GAT & 76.9 & 65.3 & 35.5 & 19.0 & 19.0 & 15.3 \\
                    & DAGNN & 72.6 & 73.8 & 76.2 & 76.4 & 76.2 & 76.0  \\
                    {\textbf{Citeseer}} & GRAND & 77.4 & 77.8 & 77.7 & 76.9 & 75.7 & 64.2  \\
                    & GCNII & 75.3 & 75.7 & 76.3 & 77.2 & 77.6 & 77.9  \\
            \cline{2-8}
                    & \textbf{DRGCN} & 78.1 & 78.4 & 78.5 & 78.5 & 78.5 & 78.7  \\
                    & \textbf{ DRGCN*} & \textbf{78.2} & \textbf{78.3} & \textbf{78.6} & \textbf{78.6} & \textbf{78.8} & \textbf{79.1} \\

            \hline
        
                    & GCN & 87.6 & 85.6 & 65.3 & 45.1 & 18.0 & 17.9\\
                    & GAT & 87.9 & 86.1 & 66.3 & 51.1 & 23.2 & 23.2 \\
                    & DAGNN & 84.8 & 85.9 & 86.0 & 84.9 & 84.0 & 82.8  \\
                    {\textbf{Pubmed}} & GRAND & 88.1 & 87.9 & 87.2 & 86.2 & 84.6 & 75.3  \\
                    & GCNII & 85.7 & 86.0 & 86.4 & 87.3 & 87.9 & 88.3  \\
            \cline{2-8}
                    & \textbf{DRGCN} & 88.6 & 89.0 & 89.1 & 89.5 & 89.3 & 89.6  \\
                    & \textbf{ DRGCN*} & \textbf{88.8} & \textbf{89.2} & \textbf{89.3} & \textbf{89.5} & \textbf{89.8} & \textbf{89.9} \\
            \hline

    \end{tabular}
    \end{table}
    \addtolength{\tabcolsep}{3pt}

    \addtolength{\tabcolsep}{-4pt}
    \begin{table}[h!]
        \centering
        \caption{\label{best_result_summary}Summary of best classification accuracy(\%) results of each model on Cora, Citeseer, and Pubmed. The number in parentheses corresponds to the depth of each model.}
        \begin{tabular}{l|c|c|c}
            \hline 
            \textbf{Model} & \textbf{Cora} & \textbf{Citeseer} & \textbf{Pubmed} \\
            
            \hline
                 GCN & $81.9 \pm 0.5$ (02) & $76.8 \pm 0.5$ (02) & $87.6 \pm 0.5$ (02) \\
                 GAT & $83.2 \pm 0.5$ (02) & $76.9 \pm 0.5$ (02) & $87.9 \pm 0.5$ (02) \\
                 DAGNN & $87.3 \pm 0.1$ (16) & $76.4 \pm 0.1$ (16) & $86.0 \pm 0.1$ (08) \\
                 GRAND & $80.9 \pm 0.1$ (08) & $77.8 \pm 0.1$ (04) & $88.1 \pm 0.1$ (02) \\
                 GCNII & $87.1 \pm 0.3$ (64) & $77.9 \pm 0.3$ (64) & $88.3 \pm 0.3$ (64) \\
            \hline
                 \textbf{DRGCN} & $87.5 \pm 0.2$ (64) & $78.7 \pm 0.2$ (64) & $89.6 \pm 0.2$ (64)  \\
                 \textbf{DRGCN*} & $\textbf{88.1} \pm 0.1$ (64) & $\textbf{79.1} \pm 0.1$ (64) & $\textbf{89.9} \pm 0.1$ (64) \\
            \hline
        \end{tabular}
    \end{table}
    \addtolength{\tabcolsep}{4pt}

    \subsection{Model Depths}\label{model_depth_section}
    
    This section will show the performance of the DRGCN compared with other state-of-the-art models as the graph neural network becomes deeper. We use the Adam SGD optimizer \cite{kingma2014adam} with a learning rate of 0.001 and early stopping with the patience of 500 epochs to train DRGCN and DRGCN*. We set L2 regularization of the convolutional layer, fully connected layer, and evolving layer to 0.01, 0.0005, and 0.005 respectively.
    
    

    \subsubsection{A Detailed Comparison with Other Models.} Table \ref{detail_result_summary} summaries the results for the deep models with various numbers of layers. We observe that on Cora, Citeseer, and Pubmed, DRGCN and DRGCN* achieve state-of-the-art performance, moreover, the performance of DRGCN and DRGCN* consistently improves as we increase the model depth. Overall, the performance of GCN and GAT drops rapidly with the increase of model depths. DAGNN and GRAND show a concave down curve with respect to model depths, which means they still suffer from over-smoothing. 
    The best hyperparameters of models will be given in detail in the appendix.

    \subsubsection{Comparison with SOTA} Table \ref{best_result_summary} is a condensed version of Table \ref{detail_result_summary}, which summarizes the optimal results of each model on three citation datasets. As can be seen from Table \ref{best_result_summary}, experimental results successfully demonstrate that DRGCN and DRGCN* achieve new state-of-the-art performance compared with baseline models. In addition, DRGCN* is the most stable model considering variances.

    \subsection{Training Set Sizes}\label{training_set section}

    The previous section significantly demonstrates the effectiveness of our proposed model when the GCN becomes deeper. In this section, we conduct experiments with different training set sizes to further demonstrate that DRGCN and DRGCN* are capable of capturing information from different scales of datasets. 
    
    The following conclusions can be drawn from Table \ref{training_size_result}. Firstly, our DRGCN performs well on relative large training set sizes.
    Secondly, DRGCN* using data augmentation performs stably and achieves the best results.
    
    
    \begin{figure}[t]
        \centering
        \includegraphics[width=0.40\textwidth]{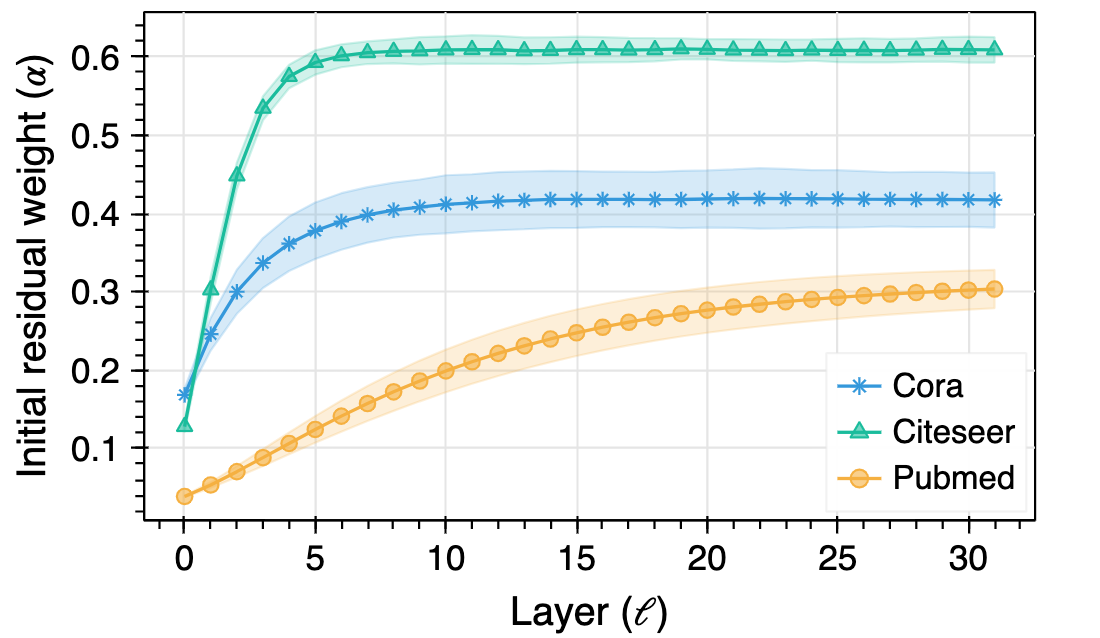}
        \caption{The evolving curve of $\boldsymbol{\alpha}$ from shallow to deep layers learned by DRGCN on Cora, Citeseer, and Pubmed. Each marker is the average $\boldsymbol{\alpha}$ of all the nodes in the corresponding layer. The shaded areas represent the 95\% confidence interval of multiple-time experiments.}
        \label{alpha_2d}
    \end{figure}

    \subsection{Ablation Study}
    \label{ablation_study section}
    
    Table \ref{ablation_result} shows the results of an ablation study that evaluates the contributions of our three techniques: the dynamic block, the evolving block, and data augmentation.
    
    The results of ``Base'' indicate that GCNII can alleviate the over-smoothing problem.
    Meanwhile, compared with "Base", the results of ``+Dyn'' shows that node personalization is beneficial for shallow layers but not enough for deep layers.
    In addition, when combining the evolving mechanism (i.e. DRGCN), the model has been significantly improved,  and this shows learning layer correlation of residual weights is important for deep GCNs. Finally, when adding data augmentation to DRGCN (i.e. DRGCN*), we achieve further improvement. 
    
    
    

    \subsection{Visualization}\label{visualization section}
    In order to show that we do achieve dynamic residual for GCN, in this section, we visually study the characteristic of residual weights learned by DRGCN from two aspects: 1) the evolving pattern of learned residual weights; 2) the distribution of learned residual weights.

    \subsubsection{The Evolving Pattern Analysis of Learned Residual Weights.} 
    
\begin{figure*}[t]
        \centering
        \includegraphics[width=0.85\textwidth]{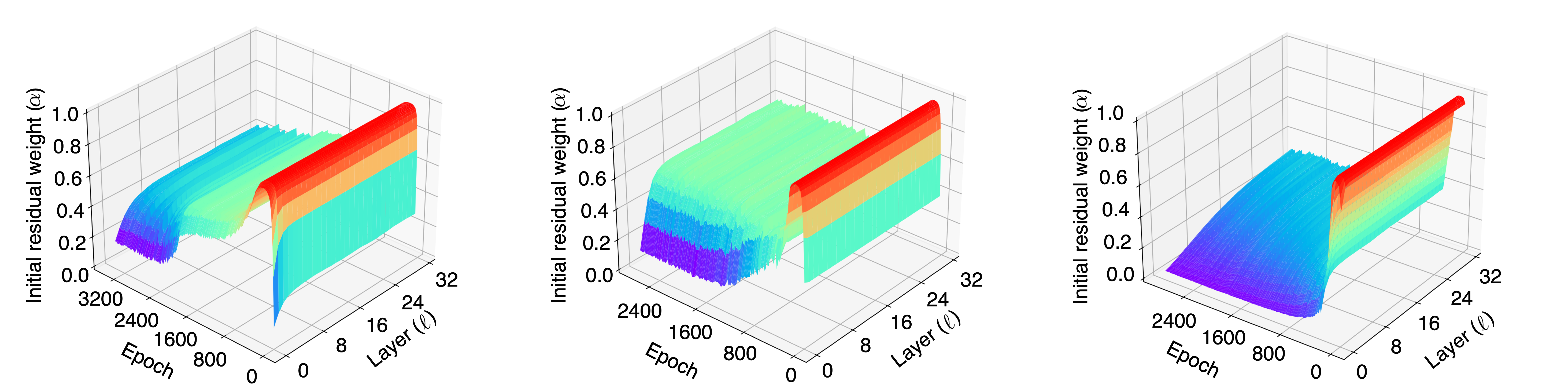}
        \caption{The convergence curve of $\boldsymbol{\alpha}$ learned by DRGCN on Cora (left), Citeseer (middle) and Pubmed (right).}
        \label{alpha_3d}
    \end{figure*}

    As shown in Figure \ref{alpha_2d}, we visualize the evolving curve about dynamic residual learned by DRGCN on Cora, Citeseer, and Pubmed. Through analysis, we make three conclusions. Firstly, with the continuous iteration of the training epoch, the residual weight $\boldsymbol{\alpha}$ in each layer converges to the curve shown in figure 3. It can be seen that $\boldsymbol{\alpha}$ is small in shallow layers, and it gradually becomes larger and converges to a constant value as the model depth increases. The overall evolving trend of multiple datasets is consistent. Secondly, the convergence value of $\boldsymbol{\alpha}$ on multiple datasets is different, which indicates that the evolving pattern of different datasets is also various. Thirdly, the inflection points of the curves are different, indicating that the boundary between short- and long-distance information is different for multiple datasets. In Figure \ref{alpha_3d}, we show the detailed convergence process of the residual weight $\boldsymbol{\alpha}$ on Cora, Citeseer, and Pubmed during the continuous training of DRGCN. Moreover, it converges eventually no matter what the initial value of $\boldsymbol{\alpha}$ is.

    \addtolength{\tabcolsep}{1pt}
    \begin{table}[t]
        \centering
        \caption{\label{training_size_result}Summary of best classification accuracy(\%) results with different training set sizes on Cora. 
        }
        \begin{tabular}{l|ccccc}
            \hline 
            \textbf{Model} & \multicolumn{5}{c}{\textbf{Training Set Sizes}} \\
            & 140 & 500 & 750 & 1000 & 1250  \\
             
            \hline
            GCN & 80.3 & 81.3 & 81.7 & 81.9 & 82.4  \\ 
            GAT & 83.1 & 83.2 & 82.7 & 83.5 & 83.7 \\ 
            DAGNN & 84.4 & 86.8 & 87.0 & 87.3 & 87.5  \\
            GRAND & 85.5 & 83.1 & 82.3 & 80.7 & 79.7 \\
            GCNII & 85.3 & 85.6 & 86.3 & 87.1 & 86.7 \\
            
            \hline
            \textbf{DRGCN} & 58.0 & 73.9 & 80.3 & 87.5 & 87.5 \\
            \textbf{DRGCN*} & \textbf{85.8} & \textbf{86.4} & \textbf{87.2} & \textbf{88.1} & \textbf{88.0} \\
            \hline
        \end{tabular}
    \end{table}

    \addtolength{\tabcolsep}{-4pt}
    \begin{table}[t]
        \centering
        \caption{\label{ablation_result}Ablation study of classification accuracy(\%) results for DRGCN and DRGCN* on Cora. ``Base'' means well-tuned GCNII. 
        ``+Dyn'' means only using the dynamic block. ``+Dyn \& Evo'' denotes DRGCN, and ``+Dyn \& Evo \& Aug'' denotes DRGCN*.}
        \begin{tabular}{l|cccccc}
            \hline 
            \textbf{Model} & \multicolumn{6}{c}{\textbf{Layers}} \\
             & 2 & 4 & 8 & 16 & 32 & 64  \\
           
            \hline
            \textbf{Base} & 84.6 & 85.6 & 86.0 & 86.4 & 86.7 & 87.1  \\
            
            \hline
            \textbf{+Dyn} & 85.4 & 86.2 & 86.4 & 86.5 & 86.6 & 86.9 \\ 
   
            \hline
            
            \textbf{+Dyn \& Evo} & 85.9 & 86.7 & 87.0 & 87.3 & 87.3 & 87.5 \\
         
            \hline
            \textbf{+Dyn \& Evo \& Aug}
             & \textbf{86.6} & \textbf{87.2} & \textbf{87.4} & \textbf{87.8} & \textbf{87.9} & \textbf{88.1} \\
            
            \hline
        \end{tabular}
    \end{table}
    \addtolength{\tabcolsep}{4pt}

    \subsubsection{The Distribution Analysis of Learned Residual Weights.} 
    
\begin{figure*}[t]
        \centering
        \includegraphics[width=0.85\textwidth]{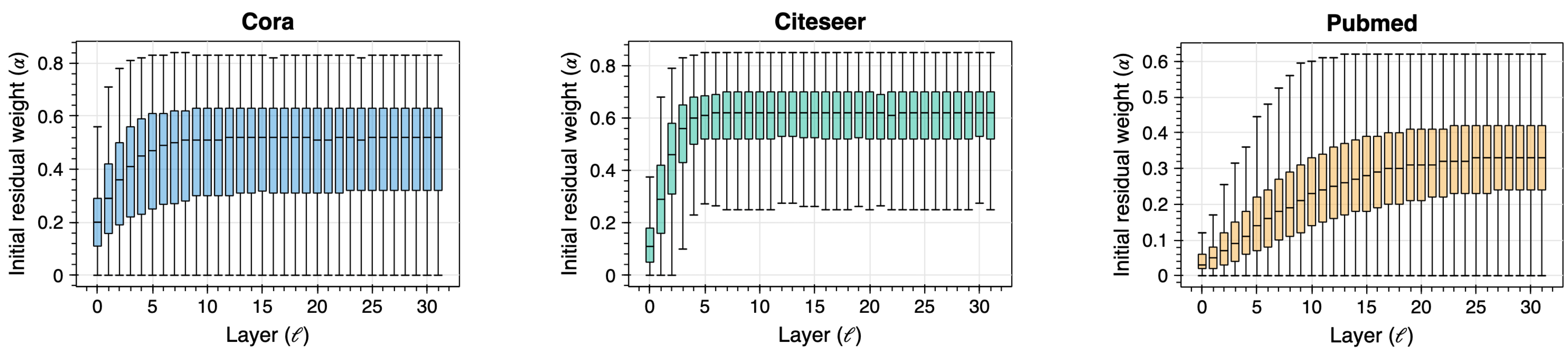}
        \caption{The quartile chart of $\boldsymbol{\alpha}$ learned by DRGCN in each layer on Cora, Citeseer, and Pubmed.}
        \label{alpha_2d_std}
    \end{figure*}

    Figure \ref{alpha_2d_std} shows the quartile chart of the residual weight $\boldsymbol{\alpha}$ in each layer after DRGCN converges. We can summarize the following three observations. Firstly, DRGCN learns personalized residual weight $\boldsymbol{\alpha}$ for each node in different layers. Secondly, the distribution trend of the residual weight $\boldsymbol{\alpha}$ also has an evolving process, and the overall trend is consistent with Figure \ref{alpha_2d}. Thirdly, in general, the distribution of $\boldsymbol{\alpha}$ is more dense and consistent in shallow layers, and as the model deepens it becomes more dispersed, which suggests that the individual differences of long-distance information on the graph are more obvious and diverse, and it is more difficult to learn.

    \subsection{Large-Scale Datasets}\label{large_scale_section}
    
    
    

    Aligning to baseline models for solid comparison on small-scale datasets, DRGCN uses the basic GCN and RNN modules. However, to improve the performance on large-scale datasets, GCN and RNN in DRGCN are upgraded to GAT and LSTM respectively, which is the DRGAT in Table \ref{ogbn_arxiv_result}. Then, we conduct experiments on ogbn-arxiv, and DRGAT achieves state-of-the-art performance compared with baseline models.
    
    Meanwhile, we propose the variant of DRGAT named DRGAT+GIANT-XRT+SelfKD which uses self-knowledge distillation \cite{zhang2019your} and new features pre-trained by GIANT-XRT \cite{chien2021node}, and it reaches SOTA result on the ogbn-arxiv dataset of the OGB leaderboard \cite{hu2020ogb} at the time of submission (team anonymous).

    
    
    

    \addtolength{\tabcolsep}{-5pt}
    \begin{table}[t]
        \centering
        \caption{\label{ogbn_arxiv_result}Summary of classification accuracy(\%) results on the ogbn-arxiv dataset compared with SOTA on OGB leaderboard.}
        \begin{tabular}{l|cccc}
            \hline 
            \textbf{Model} & \textbf{Test Acc} & \textbf{Mem} & \textbf{Params} &  \textbf{Rank} \\
           
            \hline
            
            {GCN} & $71.74 \pm 0.29$ & 1.90 & 143k & 49  \\
            
            {DAGNN} & $72.09 \pm 0.25$ & 2.40 & 43.9k & 42 \\
            
            {GCNII} & $72.74 \pm 0.16$ & 17.0 & 2.15M & 36 \\
            
            {GAT} & $73.91 \pm 0.12$ & 5.52 & 1.44M & 19 \\
            
            {RevGAT} & $74.02 \pm 0.18$ & 6.30 & 2.10M & 14 \\
            
            \textbf{DRGAT} & $\textbf{74.16} \pm \textbf{0.07}$ & 7.42 & 1.57M  & 10 \\

            \hline

            \textbf{\makecell[l]{DRGAT\\+GIANT-XRT}} & $\textbf{76.11} \pm \textbf{0.09}$ & 8.96 & 2.68M & 3 \\ 

            \hline

            \textbf{\makecell[l]{DRGAT\\+GIANT-XRT+KD}} & $\textbf{76.33} \pm \textbf{0.08}$ & 8.96 & 2.68M &  \textbf{1} \\

            \hline
        \end{tabular}
    \end{table}
    \addtolength{\tabcolsep}{5pt}

    \addtolength{\tabcolsep}{-4pt}
    \begin{table}[t]
        \centering
        \caption{\label{mini_batch_result}DRGAT-Full-Batch vs. DRGAT-Mini-Batch on the Cora, ogbn-arxiv and ogbn-products. ``DRGAT'' means DRGAT-Full-Batch, and ``DRGAT-MB'' means DRGAT-Mini-Batch. }
        \begin{tabular}{l|cc|cc|cc}
            \hline 
             
            & \multicolumn{2}{c}{\textbf{Cora}} & \multicolumn{2}{c}{\textbf{ogbn-arxiv}} & \multicolumn{2}{c}{\textbf{ogbn-products}}  \\
            
            \cline{2-7}
            
            & \textbf{Acc} & \textbf{Mem}  & \textbf{Acc} & \textbf{Mem}  & \textbf{Acc} & \textbf{Mem}   \\

            \hline
            \textbf{DRGAT} & 88.20 & 1.28 & 74.16 & 7.42  & - & -   \\
            
            \hline
            \textbf{DRGAT-MB} & 88.00 & 0.49 & 74.08 & 2.27 & 82.30 & 10.38  \\ 
            
            \hline
        \end{tabular}
    \end{table}
    \addtolength{\tabcolsep}{4pt}

    \subsubsection{Full-Batch vs. Mini-Batch Training.}
    In the previous section, DRGCN or DRGAT runs experiments with full-batch training. To adapt our model to larger-scale scenarios, we propose a memory-efficient version DRGAT-Mini-Batch. We conduct several experiments to compare the performance of DRGAT-Full-Batch and DRGAT-Mini-Batch on Cora, ogbn-arxiv and ogbn-products, and detailed experimental results from Table \ref{mini_batch_result} show that mini-batch training further greatly reduces the memory consumption of DRGAT without degrading model performance. In addition, the complete Full-Batch and Mini-Batch pseudocodes are provided in the Appendix. 
    


\section{Conclusions and Future Work}
		
	In this paper, we propose a novel deep dynamic residual GCN model named DRGCN, which can self-adaptively learn the initial residual weights for different nodes and layers. Comprehensive experiments show that our model effectively relieves the over-smoothing issue and outperforms other state-of-the-art methods on both small-scale and large-scale datasets at the same time. Visualizing the convergence and distribution of residual weights in different layers suggests that our model can capture the evolving process of the initial residual weights, which emphasizes the importance of learned residual weights and consequently has a positive effect on model performance.
	Our model can be a general technique for all the deep GCN variants incorporating the residual connection. In future work, we will conduct more experiments on different graph tasks to further verify our superior performance.

\bibliography{aaai23}

\end{document}